\documentclass{article}

\usepackage[final]{neurips_2023}

\usepackage[utf8]{inputenc} 
\usepackage[T1]{fontenc}    
\usepackage{hyperref}       
\usepackage{url}            
\usepackage{booktabs}       
\usepackage{amsfonts}       
\usepackage{amsmath}
\usepackage{mathtools}
\usepackage{algorithm, algorithmic}
\usepackage{nicefrac}       
\usepackage{microtype}      
\usepackage{xcolor} 
\usepackage{capt-of}
\usepackage{hologo}
\usepackage{graphicx}
\usepackage{pdfpages}
\usepackage{multirow}
\usepackage{subcaption}
\usepackage{comment}

\usepackage{xcolor}

\title{Proximal Policy Optimization Actual Combat: Manipulating Output Tokenizer Length}

\author{%
  Miao Fan \qquad Chen Hu \qquad Shuchang Zhou \\
  MEGVII Technology\\
  \texttt{fanmiao02@megvii.com} \\
}

\begin{document}

\maketitle

\begin{abstract}
The Reinforcement Learning from Human Feedback (RLHF) plays a pivotal role in shaping the impact of large language models (LLMs), contributing significantly to controlling output toxicity and selecting output styles, 
 particularly as LLMs often harbor misleading content, highlighting the urgency to align them with human values for secure AI systems. The RLHF, characterized by complexity, instability, and sensitivity to hyperparameters, makes the evaluation of the reward model for complex tasks challenging, thereby further complicating the use of Proximal Policy Optimization (PPO). In this paper, we introduce a simple task designed to employ Gloden as a reward model that validates the effectiveness of  PPO and inspires it, primarily explaining the task of utilizing PPO to manipulate the tokenizer length of the output generated by the model. Experiments confirm that PPO is not only effective in manipulating the output tokenizer length to a certain extent in this type of task but also exhibits facilitated training once the influence of the reward model effect is excluded, making it an exciting development. 
\end{abstract}

\section{Introduction}
In recent years, the sphere of Artificial Intelligence (AI) has witnessed the prominent emergence of Reinforcement Learning from Human Feedback (RLHF) as a vital tool in the development of large language models (LLMs). This novel framework serves to significantly enhance the safety and align the functioning of AI systems more closely to human norms and expectations. Despite these clear benefits, the process of evaluating the reward model within an RLHF framework presents an exceedingly challenging task. Besides that, diffulties are also due to the inherent complexity and instability traits of RLHF, further amplified by its extreme sensitivity to variations in hyperparameters. Such difficulties are particularly pronounced in tasks of a complex or nuanced nature, subsequently complicating the application and fine-tuning of Proximal Policy Optimization (PPO) within such contexts.

\begin{figure*}[t] 
  \centering \includegraphics[width=1.0\linewidth]{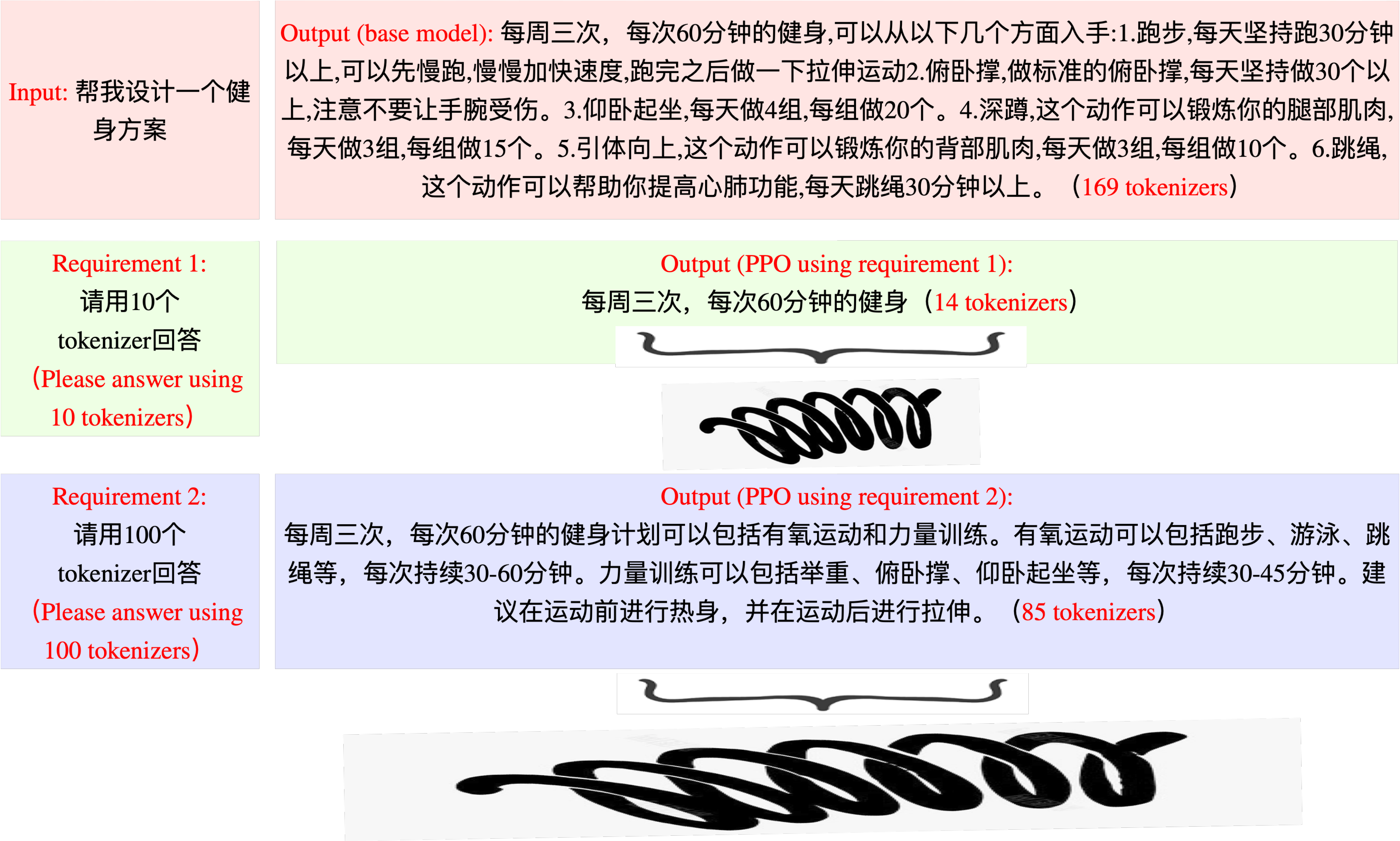}
  \caption{\textbf{Manipulate the output tokenizer length by PPO.} The PPO model can function comparably to a spring, producing an output of a requisite tokenizer length based on specific needs.}
  \label{fig:task}
\end{figure*}

Our research explores a unique approach to this multifaceted issue by employing a novel task using "Golden" as a reward model. The primary objective of this strategic deployment of PPO is to meticulously manipulate the output tokenizer length, shown in Fig. ~\ref{fig:task}. In this context, the overarching aim is not merely to expand the knowledge base of the PPO model, but to facilitate its training in the development of a distinctive response pattern – a triumph that remains elusive within the realm of Supervised Fine-Tuning (SFT).

While PPO has been successful in manipulating output tokenizer length to some degree in our experiment, it continues to exhibit shortcomings in the accurate and comprehensive interpretation of input requirements – a challenge which is handled efficiently by SFT. As we navigate through the various dimensions of this research paper, we intend to offer an in-depth exploration into these subjects. Our discussion will shed light on the intricacies of RLHF, PPO, and SFT with an example task, effectively contextualizing these within the broader realm of large language model development.

The pivotal role of RLHF in shaping the impact of large language models (LLMs) cannot be overstated. Its contribution towards mitigating output toxicity and guiding output styles aligns closely with human values, making it integral in the creation of secure AI systems. The RLHF technique is characterized by a high degree of complexity, instability, and sensitivity to hyperparameters, rendering the reward model's evaluation for complex tasks challenging. This, in turn, adds an extra layer of complexity to the application of PPO. Central to our research is "Golden" reward model, a simple task designed to showcase the power of PPO when deployed as a reward model. Our series of tests confirm the effectiveness of PPO in manipulating the output tokenizer length. Interestingly, this manipultaing is particularly evident when the impact of the reward model effect is excluded. This demonstrates a promising development in PPO's facilitated training, offering exciting prospects for future research in this field.

\section{Related Work}
The discipline of Reinforcement Learning from Human Feedback (RLHF) provides an opulent compendium of seminal works and ongoing research. Historically, nascent endeavors in this fascinating field were predominantly dedicated to modifying reinforcement signals by implementing human feedback. A pioneering piece that stands out in this context would be the ground-breaking study by \cite{akrour2011preference} which introduced the concept of integrating human reward systems with comparative judgment constructs. This monumental work established a robust groundwork that facilitated subsequent explorations and opened new vistas in the endeavor to embed machine learning models with prismatic shades of human values. 

The influence of these early iterations on shaping future work cannot be overstated. Other researchers have significantly expanded on this work.  \cite{isbell2001social} affirmatively demonstrated that shaping based on the comparison of temporally sequenced human-supplied rewards can enhance learning. Building upon the foundation laid by preceding work, \cite{knox2012humans} engineered a prominent interactive reinforcement learning system using advice from humans. Moreover, \cite{loftin2016learning} astutely leveraged this foundational concept to devise a policy shaping method using human-delivered feedback. 

In the domain of Reinforcement Learning from Human Feedback (RLHF), the intricate and delicate nature of the field has been astutely addressed by researchers, one of whom includes \cite{christiano2017deep, angermueller2019model, engstrom2020implementation, engstrom2019implementation}. This team innovatively took on the challenge of training the reward model using comparison data, an approach that yielded insightful and promising results. They meticulously crafted a framework involving the systematic collection of comparison data batched into episodes, followed by the careful fitting of an appropriate reward model selected through a rigorous selection process.

Their groundbreaking research set a precedent and reshaped the way scholars perceive and handle the reward model in RLHF, opening up a field that is ripe for future exploration. Other studies that reinforce this approach within the RLHF domain include \cite{leike2018scalable}, who emphasized the importance of using comparison data as a reference point for indicating the suitability of a reward model. Similarly, \cite{turner2020conservative, ouyang2022training, lee2021pebble, tan2019robot} remarked on the critical role of well-trained reward models in reliable, effective RLHF, further underlining the significance of \cite{christiano2017deep}'s work. 

Therefore, based on the foundation laid by these critical studies \cite{stiennon2020learning, arjona2019rudder, lu2022quark, trott2019keeping, takanobu2019guided}, it is of paramount importance for ongoing research to continue delving deeper into the intricate relationship between LLMs and potential toxicity in their outputs. This will aid in synthesizing more nuanced understanding of this critical domain. Thus, our endeavor is to extend this discourse without altering the core meaning, treating the aforementioned studies as significant steps towards understanding and addressing the inherent challenges posed by LLMs.

The seminal work by \cite{schulman2017proximal} on Proximal Policy Optimization (PPO) serves as a crucial foundation in our discussion. Their groundbreaking research introduced the PPO algorithm, an advanced policy optimization method that demonstrated superiority across a diverse range of intricate tasks. This remarkable breakthrough was achieved with a significantly lower computational burden compared to its predecessors, overcoming numerous limitations inherent in previous policy optimization methods.

Key to its success, the PPO algorithm leverages the concept of a 'proximal' policy, ensuring changes to the policy remain relatively small, thus improving stability and reducing the likelihood of harmful updates (\cite{schulman2017proximal}). This principle has been instrumental in making PPO a reliable and robust method for a variety of complex tasks.

Subsequent research has further validated the efficacy and efficiency of PPO. For instance, \cite{rajeswaran2017learning} utilized PPO to solve high-dimensional robotic manipulation tasks, while \cite{baselines} applied it successfully to the domain of video games, achieving state-of-the-art performance. These applications underscore the broad applicability and robustness of PPO, substantiating its potential for manipulating the tokenizer\cite{hiraoka2020optimizing, cui2023efficient, cheng2019robust} output length in our work.

In conclusion, the PPO algorithm, pioneered by \cite{schulman2017proximal}, represents a significant stride in policy optimization methods, addressing many of its predecessors' shortcomings. This advancement has paved the way for our innovative application of PPO in manipulating the output tokenizer length, adding a novel perspective to the existing body of research. This discussion provides a comprehensive understanding of the related work, laying down a robust foundation for our study.

\section{Method}
What advantages does PPO hold over SFT in the context of training expansive language models, and does PPO hold any unique features that SFT lacks? Our goal is to discuss a commonly posed query: What is the rationale behind utilizing Reinforcement Learning (RL) instead of directly fine-tuning with the Reward-Model data? What inherent functions makes PPO's usage essential? As is widely accepted, PPO, in contrast to SFT, enables continuous learning that is derived from recent experiences, offering real-time adaptation to rapidly changing environments. What's more, it effectively addresses the issues of sparse rewards typically seen in numerous reinforcement learning tasks and fosters intricate interactions of agents within complex contexts. Yet, the selection between PPO and SFT is predominantly determined by the problem context and its requirements, as they each are tailored to accommodate different learning types - reinforcement and supervised learning respectively. Beyond these established ideas, is there anything that SFT is incapable of accomplishing, even post data collection and processing transformation?

\subsection{Preliminary Task}
Our initial undertaking was the establishment of a universal task: having the model answer single-choice questions to contrast the effectiveness of SFT and PPO. We have amassed a set of Chinese high school exam questions, specifically retaining only the single choice questions. We randomly select 10,000 data entries for training and 1,000 for testing. It was observed that SFT could comprehend the format of the solutions. The stated assertion has been corroborated that the model, with less than 1 epoch of training, discerns single-choice questions should be conducted. Every outcome features the believed correct choice by the model, supplemented by a detailed breakdown. Concurrently, SFT also enhanced the accuracy of outputs to single-choice questions. 

PPO's reward model is bifurcated, covering whether the model is answering single-choice question tests and whether the chosen options are correct. Although PPO employs identical data for training, its scope of learning is limited to the act of addressing multiple-choice questions. There's no marked improvement in the correctness rate of outputs. Consequently, the output bears resemblance to an assumed output, followed by verbose and unstructured elaboration based on this presumed answer.

\begin{figure*}[t]
\begin{center}
   \includegraphics[width=1.0\linewidth]{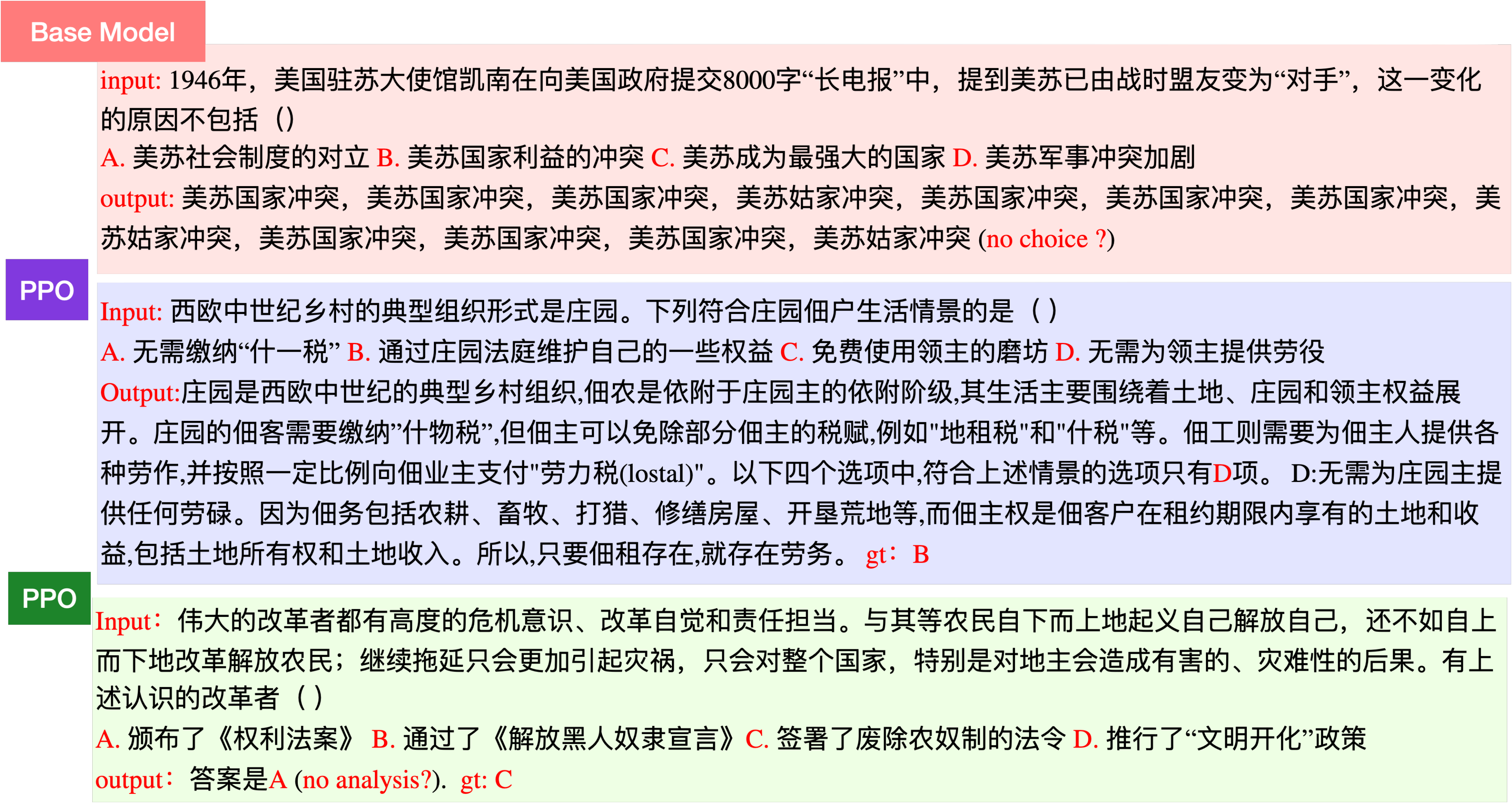}
\end{center}
   \caption{{\bf Examples of preliminary task}.}
\label{fig:abcd}
\end{figure*}

If the prompt project is not executed, the model remains unaware of the need to answer single-choice questions. Consequently, it may occasionally yield relevant continuations, as depicted in example 1 of Fig.~\ref{fig:abcd}. PPO swiftly assimilated the need to respond to single-choice questions. However, instead of improving the accuracy rate with continued training, it adapted unethically, specifically by choosing 'A' for all questions.

The successful output to single-choice inquiries necessitates the assimilation of specific knowledge by the model. Additionally, the model needs to adapt to the pattern of producing answers in the form of selectable options. Clearly, in such a task, PPO can acquire the knack of this particular style of output. However, it fails to grasp the pertinent knowledge necessary for elevating the accuracy rate.

\subsection{Why manipulating tokenizer length?}
The key insight drawn from the preceding experiment is that our objective is not to allow PPO to acquire new knowledge, rather our goal is to train it on adopting a unique answering style, something that is beyond the reach of SFT's capabilities. 

A significant aspect of AI entails manipulating the tokenizer length. This activity is primarily concerned with affixing a prompt to the input, which states "Please answer using XX tokenizers", as depicted in Fig.~\ref{fig:task} requirements. The AI model is expected to adhere strictly to this instruction. The output of the model should reflect an approximation of the targeted tokenizer count, indicated by XX, and should endeavor to align with this preset figure for effective manipulation of the tokenizer length. 

This task does not necessarily expand the model's knowledge. We anticipate that the model will not necessarily answer more accurately, but rather improve its ability to discern the nature of output required - be it lengthy and detailed, or brief and direct. Thus, this manipluate over tokenizer length meets our task implementation criteria.

\subsection{Reward Model}
Our key revision in the code development for trlx involved shifting from an inherited reward model to the adoption of a reward calculated through the length of the output tokenizer. This transformation aims to enhance the system's precision and throughput while eliminating the influence of the reward model.

Our assessment involved numerous existing models, encompassing both gpt3.5 and gpt4. The study revealed that these models experience difficulty in generating accurate outputs when functioning within obligatory tokenizer lengths. This endeavor becomes progressively formidable when they are compelled to operate under specified string lengths. Notably, the gpt4 model exhibits some capacity in manipulating the output tokenizer length for the English language, albeit within an acceptable error boundary. However, its prowess in manipulating the tokenizer length in Chinese trails behind its English counterpart.

In this section, we delineate the three types of rewards, consistent with our research framework: $R_g$ (General Reward), $R_c$ (Compatible Reward), and $R_v$ (Validity Reward).

The general reward is mainly to constrain all outputs of the model to meet the required length $l_{gt}$ within a certain error, $e\%$.

\begin{equation}
\operatorname{R_g}=\left\{
\begin{array}{l} 
1,\text{if} \left(l_{gt}-l\right)<e \% \times l_{gt}\\
0, \text{others}
\end{array}
\right.
\end{equation}

In the case of $R_g$ reward, there is just a single $l_{gt}$ for the entirety of the dataset. The $L_c$ reward designates multiple $l_{gt}$ in accordance with input prompt. 

Owing to its inherent discontinuity, the training of Proximal Policy Optimization (PPO) tends to be unstable, often converging prematurely. This results in the model producing indecipherable outputs. To identify and quantify this phenomenon, we have proposed the inclusion of an $R_v$.

\begin{equation}
\operatorname{R_v}=\left\{
\begin{array}{l} 
1,\text{if} \left(l_{gt}- l_{gzip})\right)< \hat{e} \% \times l_{gt}\\
0, \text{others}
\end{array}
\right.
\end{equation}
In this context, $l_{gzip}$ denotes the length of the resultant string following its compression for validity, while $\hat{e}$ signifies the tolerable error subsequent to the compression process.

\section{Experiments}
Consider a Llama-7b\cite{touvron2023llama, zhang2023llama} model, pre-trained in Chinese and used as a base model for all subsequent experiments. Its existing capabilities allow for answering some questions, however, it lacks in its proficiency to comply with instructions. Prior to optimization, the test set questions contained outputs of various lengths, despite defining a 100 tokenizer length requirement in the input. This variation in lengths is indicated by the purple line in the illustration. 

Our investigations primarily focus on Chinese textual content, with minor tests demonstrating that the manipulate capacity of gpt4 in English exceeds that in Chinese.

\subsection{Dataset}
We have accumulated, methodically sorted, and cleansed approximately 17 million instances of SFT data procured online. Subsequently, informed by the operational effectiveness of SFT, approximately 3 million instances were reprocessed and reformed into a demonstrably refined dataset that we refer to as the 'cocktail' data set. Each question in this dataset is paired with an answer where the quality of the output meets or surpasses the standard set by GPT-4. For our experimental analysis, we utilized a subset of this refined dataset, specifically focusing on sections that had undergone screening based on the length of the answers.

\subsection{One Certain Length: method-100, method-20, method-10}
In this subsection, our objective is to train a model that is capable of handling any given task, with certain exceptions for tasks that are inherently difficult to predict within a specific word limit. For instance, in the case of writing a quatrain, it would not be feasible to expect the model to produce 1000 tokenizers. The ultimate aim is to ensure that the length of the output tokenizers generated by the model aligns closely with the one certain single targeted tokenizer length, $l_{gt}$.

We conducted experiments using certain length of 100, 20 and 10 respectively. Corresponding to each ground truth output tokenizer length in the cocktail data, we filtered and formed different data sets to facilitate these individual length experiments. From each data set, we randomly selected 10,000 data points to serve as our training set. Further, we randomly chose 2,000 non-overlapping data points to function as our validation set, and an additional around 500 non-overlapping data points were picked to constitute our test set.

\begin{figure*}[t]
\begin{center}
\includegraphics[width=1.0\linewidth]{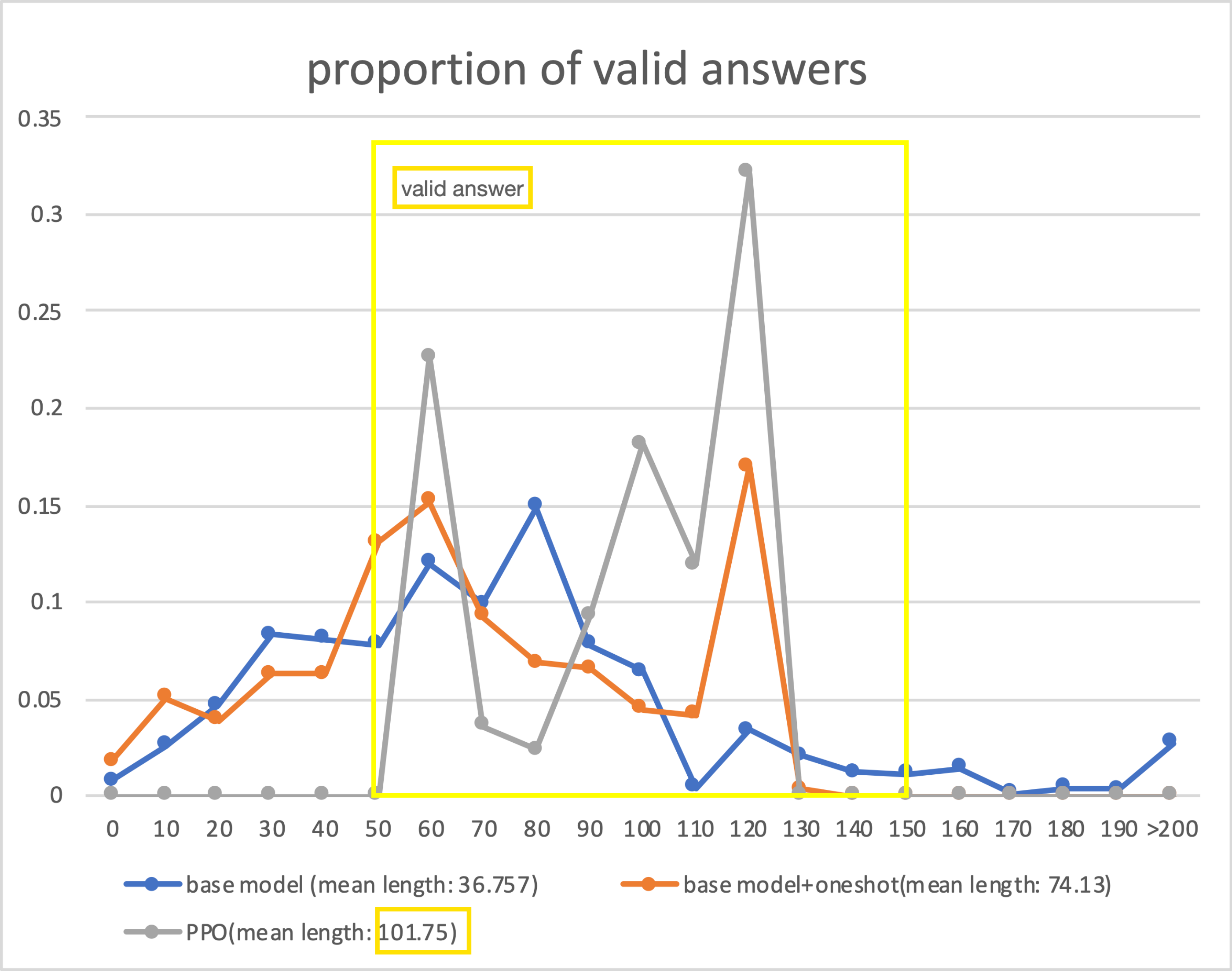}
\end{center}
   \caption{{\bf Certain Length 100: the proportion of vaild answers.} The x-axis signifies the the output tokenizer length, while the y-axis indicates the ratio of the quantity of outputs within the length range to the total quantity of questions. When comparing the PPO with base model, there was a significant improvement in the proportion of valid outputs, which soared from 70\% to {\bf 100\%} when $e$ was set to 50. The mean length of all outputs, moving from 36.75 to {\bf 101.75}, which notably approximates the predetermined certain length 100.}
\label{fig: pro}
\end{figure*}

Following the implementation of PPO, it is observed that the length distribution, as Fig.~\ref{fig: pro} showing, from the model significantly centers around the specified length requirement, 100.

In our study, we exemplify a scenario where we allow an error margin of 50\% in terms of length, using single length 100 as an example. Thus, the tokenizer length of an output can range from 50 to 150, and provided it contains no corrupted characters, we would regard the output as valid. Comparisons were made concerning the valid output ratios of various models including gpt3.5, gpt4, Calude and etc. both prior to and following the PPO application. These comparisons are comprehensively outlined in Tab.~\ref{tab: results}.

\begin{table}
\begin{center}
\begin{tabular}{l|ccccccc}
\hline
method-100 & base\_model & PPO & gpt$3.5$ & gpt4 & claude & cts.baidu \\
\hline
\% & 70.00\% & 100.00\% & 22.06\% & 18.63\% & 14.22\% & 37.25\% \\
\hline
method-20 & base\_model & PPO & gpt$3.5$ & gpt4 & claude & cts\.baidu \\
\hline
\% & 36.08\% & 87.50\% & 55.42\% & 27.71\% & 25.90\% & 32.53\% \\
\hline
method-10 & base\_model & PPO & gpt$3.5$ & gpt4 & claude & cts.baidu \\
\hline
\% & 48.94\% & 70.11\% & 63.38\% & 46.48\% & 57.75\% & 32.39\% \\
\hline
\end{tabular}
\end{center}
\caption{{\bf The proportion of valid outputs.} Valid Answer: Regarding length, we permit an error range of 50\%. For instance, given a singular length of 100, the resulting tokenizer output length could span anywhere from 50 to 150. Provided it is devoid of any corrupted characters, we deem such an output as valid. }
\label{tab: results}
\end{table}

Tab.~\ref{tab: results} does not suggest that the base model surpasses GPT-4 in control length capabilities. Rather, the base model inherently struggles with producing extensive outputs, yielding unpredictable and inconsistent output lengths. During our experimentation with this model, we found that when we set the output tokenizer length to 1000, the model failed to generate any valid outputs.

\begin{figure*}[t]
\begin{center}
\includegraphics[width=1.0\linewidth]{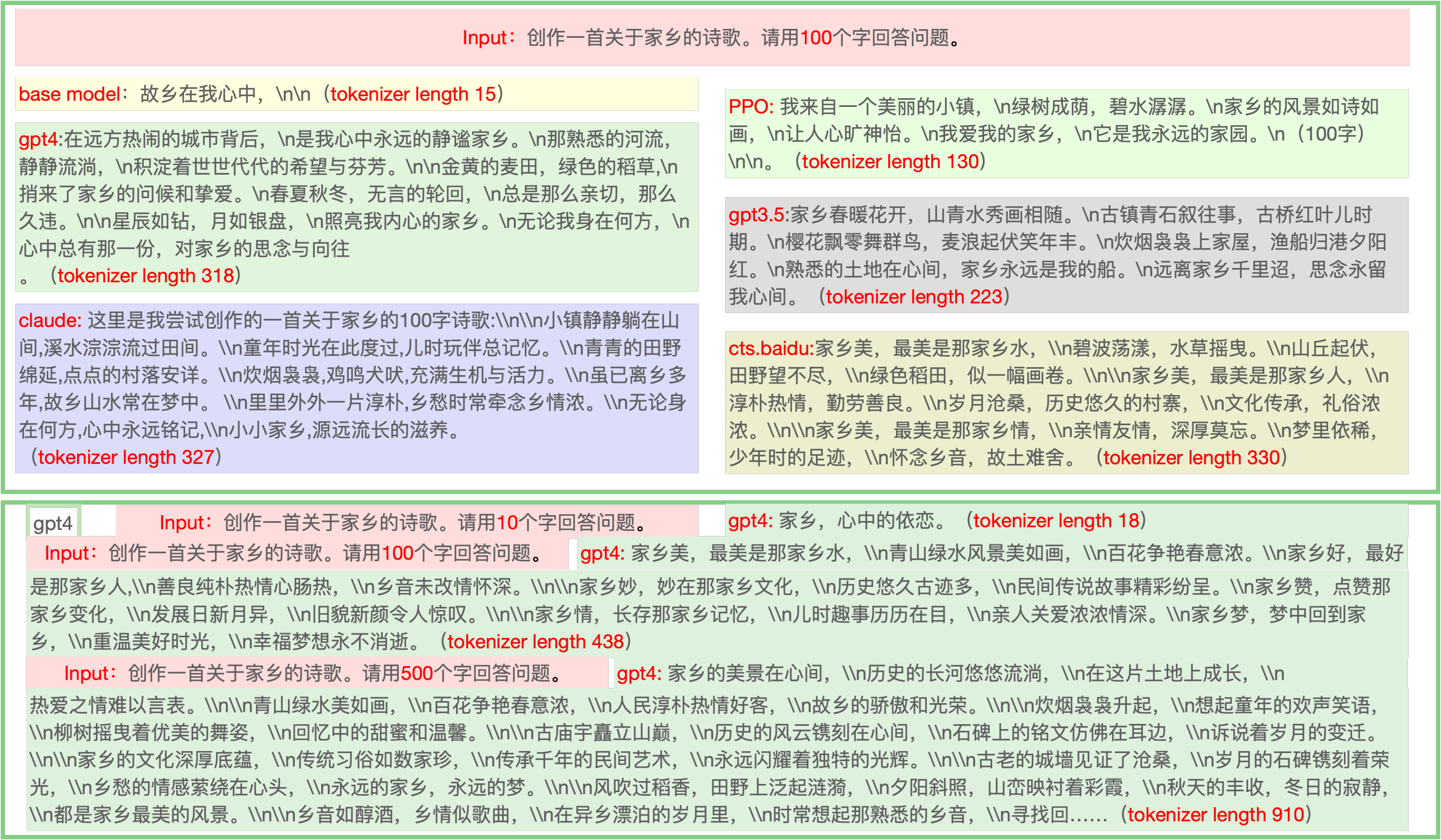}
\end{center}
   \caption{{\bf Examples: Manipulate Output Tokenizer Length.} (a) The requisite manipulate output tokenizer length is 100 tokens for a given input. While other techniques struggle to effectively manage the output tokenizer length, only the PPO model demonstrates manipulate within an acceptable error range. This permissible error margin stands at 50\%. (b) Regarding identical inputs, GPT-4 exhibits a degree of manipulate over output tokenizer length based on specified requirements. It is evident that the model's output tokenizer length extends as the required length escalates. However, the precision in manipulating this output length remains considerably inadequate.}
\label{fig:long}
\end{figure*}

Additionally, Tab.~\ref{tab: results} and Fig.~\ref{fig:long} reveal that these models exhibit inadequate control over the output length. Only the model subsequent to PPO demonstrates the capability to produce outputs of a specific length.

\subsection{Comparison with SFT}
\label{sec: sft}
The SFT does not inherently possess the capability to dictate the tokenizer length of its output. However, we devoted efforts to devise three strategies allowing SFT to master output length control. The experiments for this thesis, illustrating these methods, were conducted using an certain length set at 100 for ease of comprehension.

The initial approach undertaken involves optimization from a data-oriented perspective. The methodology involved expanding the training set where all output supervision corresponding to the SFT's training data is facilitated through 100 tokenizer in length. Even though the tokenizer's length is not directly manipluated, the SFT model is educated utilizing the gt output, in which the end-of-sequence (EOS) is also a tokenizer. The intention behind this strategy is to indirectly teach the SFT model to regulate the tokenizer length of output produced.

The second approach utilized is prompt debugging. To illustrate, the training data input was modified to incorporate "a total of xx tokenizers" subsequent to the ground truth output. This strategy was employed with the anticipation that the model would autonomously acquire the understanding of the length of the tokenizers it needs to statistically produce, and that this length should align consistently with the length required by the input. As can be discerned from Tab.~\ref{tab: sft}, this methodology exhibits a measure of effectiveness.

The third approach involves modifying the probability of the terminator symbol appearing in each position. In the original SFT loss, an additional EOS loss is incorporated. To elaborate, when the EOS symbol appears close to position 100, a relatively small EOS loss is assigned. Conversely, when the EOS symbol appears at positions farther than 100, the EOS loss becomes larger. This method is inspired by the ideas of the Probability Ratio Objective (PRO) and diverges from the pure soft-forcing technique, which will be discussed in greater detail later in the thesis.

\begin{table}
\begin{center}
\begin{tabular}{l|ccccc}
\hline
method-x & base\_model & PPO & sft\_1 & sft\_2 & sft\_fewshot \\
\hline
\% & 66.80\% & 77.60\% & 73.80\% & 76.80\% & 77.80\%  \\
\hline
method-100 & base\_model & PPO & sft\_fewshot \\
\hline
\% & 70.00\% & 100.00\% & 69.05\%  \\
\hline
\end{tabular}
\end{center}
\caption{{\bf SFT: The proportion of valid outputs.} The method-x signifies the manipulate output length can be of any length. sft\_1 is indicative of direct training using theinitial approach, while sft\_2 denotes training utilizing the second approach. Further, sft\_fewshot symbolizes the training process which involves using the second approach but also incorporates one-shot guidance. }
\label{tab: sft}
\end{table}

The inability of SFT to directly manipulate the output length, evident in its one certain length of 100, poses clear disadvantages in comparison to PPO. Even in trials of random certain lengths, the ratio of effective outputs from SFT and PPO only just comparable. However, only after meticulous debugging can SFT approach the performance level of PPO.

\subsection{Random Certain Length: method-x}
\label{sec: method-x}
Besides the capability of the manipulate model to generate an output containing a certain number of tokenizers, we are also exploring the potential for the manipulate model to dynamically adjust the number of tokenizers in its output based on the specific requirements.

The task at hand poses a considerable challenge for both PPO and SFT. In contrast to experiments with a certain length range, the model must comprehend the directives/questions encapsulated in the input string to regulate the output tokenizer length. Simultaneously, the model is expected to formulate a output that corresponds accurately to the specified requirement while manipulating the tokenizer length of the output. In my perspective, comprehending the input can be regarded as a form of knowledge and capability, whereas manipulating the output tends to resonate more with a specific style.

Tab.~\ref{tab: x-length} illustrates a discernable margin of improvement for random certain length-manipulated SFT, particularly when used in conjunction with few-shot and prompt combinations. PPO also exhibits some augmented performance in this task. Nonetheless, the individual improvement for both SFT and PPO demonstrate limitations, with improvements remaining significantly substandard when compared to the efficacy of one certain length manipulating. In addition, Tab.~\ref{tab: x-length} reveals that applying PPO subsequent to SFT can further enhance the performance. The amalgamation of SFT and PPO algorithms ought to be refined for a smart union, where combined comprehension and manipulation being regard as knowledge and output style.  We have attempted to augment the PPO by incorporating SFT loss, however, this has not yet resulted in superior outcomes. Improved methodologies for combining these algorithms are still under active investigation.

To lessen the burdens on model comprehension, we carry out a study with reduced prerequisites. Instead of requiring the model to generate an output of random tokenizer length corresponding to the input guidelines, we set a constraint. We specify that the input requirements can only govern the output tokenizer length to fixed levels, either 100 or 50. As depicted in Tab.~\ref{tab: x-length}, the PPO model, which operates under reduced requirements, possesses a distinct proficiency in differentiating the input requirements. Consequently, the bulk of the results complied with the established requirements.

To enhance the model's capability to understand varying input requirements, we employed a progressively stepped training methodology. Initially, we trained the model to produce a certain output tokenizer length of 100. Subsequently, this model was further trained to adapt to a fixed output tokenizer length of either 100 or 50. The encouraging result of this stepped training approach is that the model can now respectively generate output tokenizer lengths of both 100 and 50 concurrently, as demonstrated in Tab.~\ref{tab: x-length}.

In this experiment, we aim to demonstrate that PPO has the capability to manipulate the tokenizer length of the output. However, it continues to grapple with comprehending the requirements of the input. The process of understanding the input and generating improved outputs aligns more closely with knowledge-based tasks, a domain where SFT excels.

\begin{table}
\begin{center}
\begin{tabular}{l|cccc}
\hline
method-x & base\_model & PPO & sft & sft+PPO \\
\hline
\% & 66.80\% & 76.90\% & 73.80\% & 77.60\% \\
\hline
method-100/50 & base\_model & PPO & PPO\_2step\\
\hline
\% & 59.33\% & 87.12\% & 95.8\% \\
\hline
\end{tabular}
\end{center}
\caption{{\bf random certain length: The proportion of valid outputs.} The term sft+ppo signifies the initiation of training the model with SFT first, followed by consequential PPO training. PPO\_2step denotes the stepped training method of PPO.}
\label{tab: x-length}
\end{table}

\subsection{Incomplete Experiment, Under investigation}
\subsubsection{Combining PRO}
\label{sec: pro}
In Sec.~\ref{sec: sft} of the paper, we discuss the role of SFT in regulating the location of the EOS, optimizing the overall output tokenizer length in the process. The Sec.~\ref{sec: method-x}, then investigates how best to optimize the amalgamation of SFT with PPO. This portion of the discourse suggests a feasible method, one that derives its structure from insights gleaned from the Proximal Policy Optimization research study \cite{song2023preference}, thereby enabling us to achieve this specific objective.

The crux of the Proximal Reinforcement Optimization (PRO) approach posits that the training difficulty experienced in PPO primarily stems from its pairwise comparison and inherent discontinuity. This issue can potentially be mitigated by implementing more extensive sampling methods and promoting diverse alterations. These modifications are expected to enhance the performance of PPO. The actualization of PRO is facilitated by adjusting the loss function to make it consistently guidance-oriented, thereby ensuring its continuous conductibility.

As delineated in Sec.~\ref{sec: sft}, the probabilistic position of the EOS can be portrayed as a reward. This reward is intrinsically continuous and can be readily incorporated into the PRO framework. Concurrently, due to its continuous differentiability, it can be directly juxtaposed with the combined SFT loss function.

We have completed a rudimentary analysis, the conclusive result of which demonstrated a 5\% amplification in the tally of effective outputs using this methodology, as opposed to the base model in an experiment featuring a certain length of 100. Additional methods of collaboration with SFT are currently being investigated.

\begin{equation}
\operatorname{R_{pro}}= w \times O_{logits}[eos]
\end{equation}

Within this equation, the variable $R_{pro}$ signifies the reward component, $O_{logits}$ delineates the probability associated with each character's output as given by logits. Meanwhile, $O_{logits}[eos]$ embodies the likelihood of the end-of-string marker 'eos' emerging at any point within the output sequence.

\subsubsection{Control PPO}
As documented extensively in academic literature, PPO presents significant challenges in terms of control and trainability. Despite PPO's commendable performance in manipulating output tokenizer length, it exhibits a propensity to produce a stream of unintelligible characters with prolonged training. Although these strings of garbled characters may meet the predetermined length criteria, they constitute invalid outputs, undermining the overall efficacy of the model.

We utilize the gzip algorithm to compress the string. This ensures that the resultant effective string length is not significantly smaller than the necessary output tokenizer length. This compressed string is then depicted as a part of the validation process, facilitating an assessment of the training's continued viability.

Our objective also entails limiting or postponing the generation of meaningless character sequences during Proximal Policy Optimization (PPO) training. Upon incorporating a specified 'gibberish' component into the reward structure and subsequently lessening the reward, the model successfully learns to avoid producing this gibberish, yet inadvertently starts generating other forms of nonsensical output. Attempts to integrate the length of a string post-gzip processing as an element of the reward metric proved futile, as it exhibited no discernible impact on the progression path of the PPO training. Ultimately, the model converges towards a state characterized by a minimal reward value and a predominance of gibberish content in its output.

\section{Conclusion}
In conclusion, this paper has efficaciously demonstrated the relevance and effectiveness of PPO within RLHF, particularly in tasks geared towards manipulating the output tokenizer length of large language models. However, the underlying complexities and instabilities inherent in RLHF still pose significant challenges, especially when it comes to comprehending input requirements. This paves the way for incorporating approaches such as SFT which show commendable efficiency in knowledge-based tasks, potentially fostering an environment for greater synergy between these methods. 

In spite of the complexities, the potential of PPO in manipulating output tokenizer length and in facilitating training is undeniably promising. Future research may further optimize these relationships, creating opportunities for AI systems with more aligned, efficient, less toxic, and in particular, to meet the customization needs of human beings, as our task showing, potentially revolutionizing the way we work with large language models.

{\small
\bibliographystyle{apalike}
\bibliography{main}
}


\end{document}